\documentclass[letterpaper]{article} 
\usepackage{aaai24}  
\usepackage{times}  
\usepackage{helvet}  
\usepackage{courier}  
\usepackage[hyphens]{url}  
\usepackage{graphicx} 
\usepackage{colortbl}
\usepackage{natbib}  
\usepackage{caption} 
\usepackage{algorithm}
\usepackage{algorithmic}

\usepackage{bmpsize}
\usepackage{newfloat}
\usepackage{listings}

\usepackage[utf8]{inputenc} 
\usepackage[T1]{fontenc}    
\usepackage{booktabs}       
\usepackage{amsfonts}       
\usepackage{nicefrac}       
\usepackage{microtype}      
\usepackage{xcolor}         
\usepackage{xspace}
\usepackage{enumitem}
\usepackage{xcolor}
\usepackage{gensymb}
\usepackage{amsmath}
\usepackage{multirow}
\usepackage{hhline}
\usepackage{tabularray}
\usepackage{pdfpages}

\newcommand{\prjname}{DeepCompass\xspace}

\DeclareCaptionStyle{ruled}{labelfont=normalfont,labelsep=colon,strut=off} 
\floatstyle{ruled}
\newfloat{listing}{tb}{lst}{}
\floatname{listing}{Listing}
\definecolor{lightgray}{gray}{0.9}

\title{\prjname: AI-driven Location-Orientation Synchronization\\ for Navigating Platforms}
\author{
    Jihun Lee, SP Choi, Bumsoo Kang\textsuperscript{\rm *}
    , Hyekyoung Seok, Hyoungseok Ahn, Sanghee Jung
    }
\affiliations{
    Lotte Mobility AI Research\\
    \{jihoon8798, seungpyo.choi, hyekyoung\_seok, hyeongseok\_ahn, sanghee.jung\}@lotte.net, \textsuperscript{\rm *}steve.kang@kaist.ac.kr
}

\usepackage{bibentry}

\begin{document}

\maketitle

\begin{abstract}
In current navigating platforms, the user's orientation is typically estimated based on the difference between two consecutive locations. In other words, the orientation cannot be identified until the second location is taken.
This asynchronous location-orientation identification often leads to our real-life question: \textit{Why does my navigator tell the wrong direction of my car at the beginning?} 
We propose \prjname to identify the user's orientation by bridging the gap between the street-view and the user-view images. First, we explore suitable model architectures and design corresponding input configuration. Second, we demonstrate artificial transformation techniques (e.g., style transfer and road segmentation) to minimize the disparity between the street-view and the user's real-time experience.
We evaluate \prjname with extensive evaluation in various driving conditions. \prjname does not require additional hardware and is also not susceptible to external interference, in contrast to magnetometer-based navigator. This highlights the potential of \prjname as an add-on to existing sensor-based orientation detection methods.
\end{abstract}

\section{Introduction}
Recent advances in path planning techniques with sensor-based localization introduce various navigation services in the mobility field. 
However, they rely on sensing location changes to detect the user's orientation, and subsequently guide them on the right path to take ~\cite{almazan2013full,constandache2010towards}.
This indicates that the orientation cannot be immediately determined until the location change is detected, namely \textit{asynchronous location-orientation identification}. Since the orientation is computed based on two consecutive locations, it is unable to identify until the second location is taken. 
Smart devices are commonly used for navigation and are equipped with magnetometer.
While magnetometer in smart device could be an alternative option to detect orientation without sensing location changes, it could be unreliable in certain situations such as surrounding magnetic field. 

\begin{figure}
    \centering
    \includegraphics[width=0.63\linewidth]{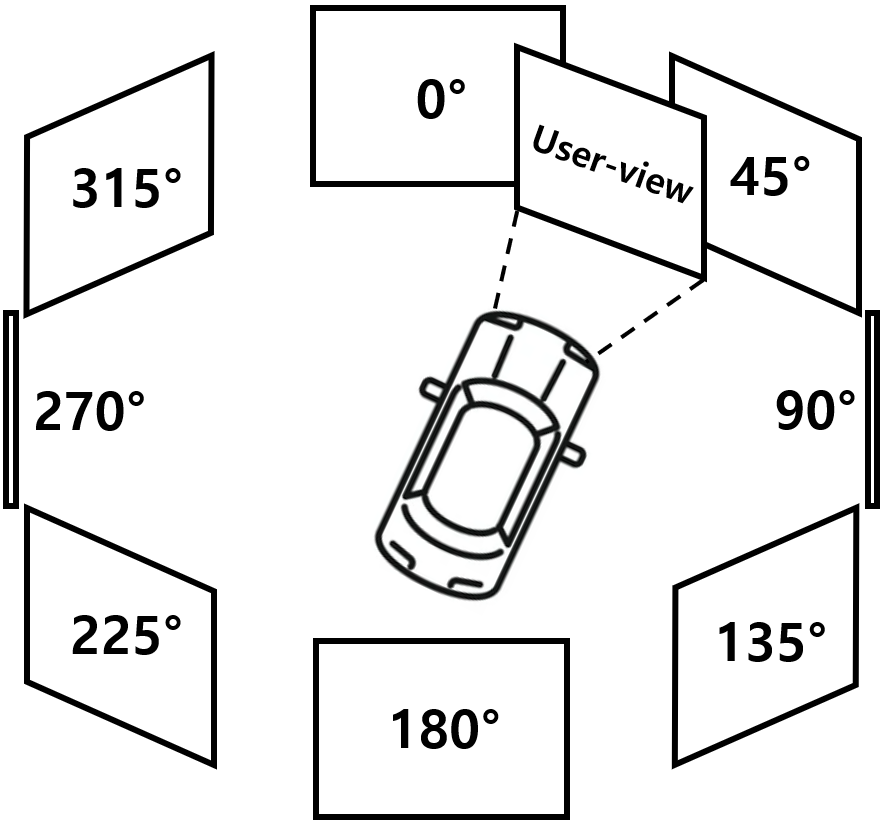}
    \caption{\prjname identifies the orientation of user (user-view) based on sliced street-views with a certain angular interval.}
    \label{fig:concept}
\end{figure}

Asynchronous location-orientation identification leads to subtle yet widespread inconvenience in our daily lives, especially when drivers require guidance from a navigator at departure. Think of Alex rushing out of the underground 
garage for her next client meeting. As Alex exits the garage, the GPS is detected and the navigator begins to recognize her location.
Due to delayed GPS recognition underground, the second location has not yet been detected, while the navigator fails to
determine which orientation Alex is heading.
However, Alex has to make a decision in a few seconds, which can lead to incorrect left/right turn guidance.
If Alex takes a wrong turn, after a while the navigator will eventually give her direction again, saying ``\textit{recalculating the route}''.
Consequently, she will have to go around a long way to get back on the right track.

In this paper, we tackle asynchronous location-orientation identification by utilizing vision deep learning techniques. 
As for the case study, we target situations when users drive ground vehicles and propose \prjname leveraging street-view and the user-view images to identify the correct orientation. We aim to correctly inference front-facing images from user's perspective (namely user-view) with a model only trained on street-view image, as labeled user-view data is insufficient. Using a camera image with vision model does not 1) suffer from external interference and, 2) require additional hardware devices, making it cost-effective when plugged on top of existing sensor-based technologies. 

The key challenges are as follows: the problem that the visual commonality is absent among the images in same label, and that the street-view used in training and the user-view used for inferencing is completely different.
In the case of a typical image classification supervised learning task, images with the same label share common visual features. However, since the commonality between images with same label in our task is an ’angle’, cannot be classified by the image itself.
Also, common machine learning models are trained using labeled data sampled from the rather similar environment in which they are expected to operate. However, our approach involves training the model using data collected from a street-view and then applying it to identify the orientation of user-view, which is in a significantly different environment (e.g., 360\degree camera and typically used camera). In addition, there are also problems regarding temporal difference, as street-view and user-view images are captured in different moment of time. This leads to a visual gap between them in terms of weather conditions, illumination (day/night), and the presence of dynamic objects (e.g., people, cars, etc.).

Figure~\ref{fig:concept} shows the overall concept of \prjname. It slices a 360\degree panoramic image of street-view into multiple flat images at a certain angle. \prjname determines the position of a user-view in the list of the sliced street-views. This approach of using multiple images is not suitable for most of the vision deep learning models as they are designed to process a single image. Additionally, in contrast to typical classification task, labels cannot be inferred by the images themselves in our situation, since there is no visual commonality among the images in the same label. Therefore, we generate input data by combining image frames containing both sliced street-view and user-view images in various ways. We then compare the performance of various deep learning models to find the most suitable input format and model structure for \prjname.

We also demonstrated \prjname in a real-world setting and showed 63.9\% accuracy, which achieve approximately 73\% of our upper bound accuracy (87.44\%). For the upper bound, we used the accuracy of models on the test data composed of only street-view images, as this accuracy can be considered as the upper limit for the models used in our experiments.




\section{Related Work}
\subsection{Visual Place Recognition (VPR)} 
VPR is a task of matching images of places with different views of the same place but taken at different times, requiring robustness on changes in various environmental factors, e.g., lighting, season. Promoting robustness to various environmental changes is key factor to recent VPR research. ~\cite{li2021unsupervised} raised the issue of place recognition performance degradation for continuous environmental changes in the same place, and suggests a solution using unsupervised learning. ~\cite{chen2023Place} proposed three ways to encode and reconstruct place images to cope with occlusion and lighting changes. ~\cite{anoosheh2019night} and ~\cite{liu2020day} used image synthesis and inpainting techniques to solve problems caused by illumination and occlusion. However, the limitation of these studies is that the orientation cannot be predicted.

Cross-view place recognition is the task of recognizing a location using a ground image and an aerial image of that location. This task even involves predicting the location and orientation of the ground camera within the aerial image. For instance, ~\cite{sarlin2023orienternet} uses 2D maps to predict the position and orientation of the camera. ~\cite{lentsch2023slicematch} focuses on the estimation of the camera pose between aerial and ground images. However, these studies have limitations in terms of their robustness to changes in the environment, such as day/night variations and seasonal differences.

\prjname does not recognize locations. It is rather an orientation detection technique that is robust to the limitations of mentioned VPR studies. \prjname determine the direction using user-view and street-view images at the certain location, assuming that the location has been correctly identified. By leveraging publicly available and easily accessible street-view images, we have enhanced the practicality. With orientation detection, we want to solve the problem of asynchronous location/direction identification that is common in navigation platforms. Table \ref{tab:comparison} summarizes differences between the existing VPR studies and \prjname.

\begin{table}[]
\centering
\small
\begin{tabular}{|c|c|c|c|}
\hline                                                      
& \begin{tabular}[c]{@{}c@{}}Orientation\\ Recognition\end{tabular} & 
\begin{tabular}[c]{@{}c@{}}Robustness to \\Environment    \end{tabular} & 
\begin{tabular}[c]{@{}c@{}}Place\\ Recognition\end{tabular} \\ 
\hline
\cellcolor{lightgray}\begin{tabular}[c]{@{}c@{}}\textbf{Deep}\\ \textbf{Compass}\end{tabular} & O & O & X \\ 
\hline
\begin{tabular}[c]{@{}c@{}}Cross-view\\ VPR\end{tabular} & O & X & O \\ 
\hline
\begin{tabular}[c]{@{}c@{}}General\\ VPR\end{tabular} & X  & O  & O \\ 
\hline
\end{tabular}
\caption{Comparison with different approaches of VPR}
\label{tab:comparison}
\end{table}

\subsection{Magnetometer Calibration}
The most widely used method for orientation detection is the magnetometer \cite{wiki:Magnetometer}. While magnetometers are generally used as a compass to detect orientations, deviation caused by metallic objects tends to make magnetometer-equipped system unreliable. Specifically, those embedded in mobile devices (e.g., smartphone) are prone to common magnetic disturbance in surrounding environments. Thus calibration becomes an necessary process. Calibration techniques have evolved in several ways, which include utilizing external equipment~\cite{gebre2006calibration} and smart-device only. The latter can also be divided into magnetometer data-exclusive calibration~\cite{fang2011novel, vasconcelos2011geometric, wu2015calibration} 
and fusion with other inertial sensors (e.g., gyroscope and accelerometer)~\cite{li2012new, kok2016magnetometer, wu2017dynamic, 10.1145/3210240.3210348}. But heading error due to fundamental magnetic field distortion remains, making it difficult to proactively deal with the unpredictable environmental changes in the real-world. Our study aims to develop a more tolerant orientation detection technique by operating independently with magnetometer or other sensors. Since a camera is used as a source for \prjname, it is basically magnetic disturbance-free.

\section{Technical Approaches}

\begin{figure*}
    \centering
    \includegraphics[width=0.8\textwidth]{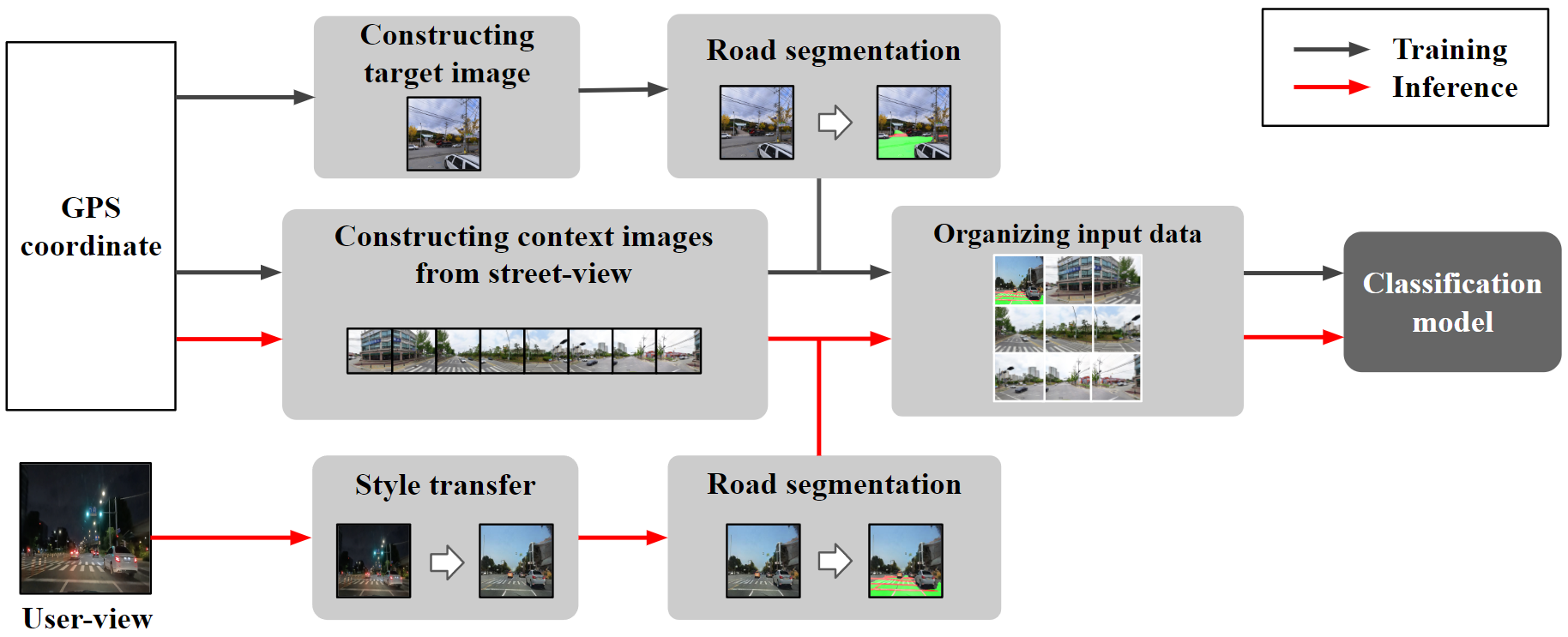}
    \caption{Architectural pipeline of the \prjname.}
    \label{fig:overview}
\end{figure*}

In this section, we outline challenges faced during \prjname implementation and our solutions. 
Figure~\ref{fig:overview} illustrates the architectural pipeline of \prjname utilizing street-view images to determine user orientation.
We explored three methods to bridge the gap between street-views and user-views. 
In street-view data with orientation as a label, images with the same label (e.g., north) mostly have no visual commonalities between them (due to the different locations). However, even in this case, it is necessary to correctly classify orientation in vision-based orientation detection. We have named this challenge the \textit{non-self-explanatory image classification problem} and aim to explore solutions for it.

\begin{figure*}
    \centering
    \includegraphics[width=0.95\textwidth]{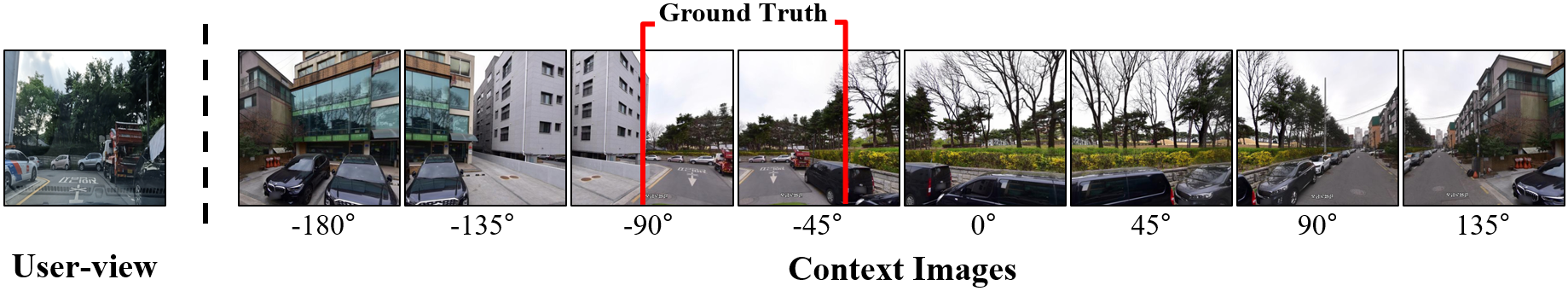}
    \caption{Example of the target image (user-view) and the corresponding context images (sliced-views). Ground truth locations in the context images are aligned with the orientation of the target image.}
    \label{fig:inputexample}
\end{figure*}

Although a number of existing work on orientation detection utilize BEV (Bird's Eye View) or 2D maps ~\cite{sarlin2023orienternet,lentsch2023slicematch}, we attempt to use street-view in our study. The advantage of using street-view is its convenience of building up-to-date data since street-view is easily accessible and allows for the continuous acquisition of the latest images.
\prjname extracts sliced images from street-views to determine the orientation of the user-view image. The sliced images, arranged in order of angle, are used as context images. The orientation of the target image, which is the user-view, is predicted by finding its position among the listed context images. Refer to Figure~\ref{fig:inputexample} for a example of the target and context images.



\subsection{Discrepancy between street-view and user-view}
The challenge lies in the differences between street-views and user-view. \prjname is trained with street-view images, but it should operate in the user-view, which has clear differences in various conditions (e.g., view point, weather, illumination). To address this challenge, we adopted three primary strategies: 1) street-views captured at different moments, 2) style transfer and 3) road segmentation. The followings are detailed descriptions of each approach.

\subsubsection{Street-views in different moments}
We use multiple street-views captured at the same location in different moments to address the issue that street-view and user-view are always different in time.
One of the advantages in utilizing street-view data is that they offer street-view at the certain location captured from various time points, ranging from the past to the present. The variation in the moments does not merely include difference in time, but also the difference in overall appearance of images, such as seasonality, illumination, or surrounding objects.
Additionally, upon attentive examination of the street-views, it was discovered that the captured locations of street-views taken at different moments did not align precisely. Thus, the variation in the moments of street views also includes changes in the capturing locations.
The discrepancy between street-view images of different time points are similar to those we aim to address in this study, which is the discrepancy between street-view and the user-view images. 

\subsubsection{Style transfer}
Since most street-view data is collected during the daytime on sunny days, solely relying on biased data for training may lead to model over-fitted on specific conditions. This makes it difficult to capture diverse range of weather and illumination depicted on the user-view images. 
To overcome this over-fitting issue, we apply style transfer techniques. 
An intuitive way of applying style transfer is to transform the training data into various environments (e.g., rainy) and train the classification model.
Reversely, we transform target images into a street-view-like representation, generating them to reflect visual attributes of daytime, sunny days. 
This approach allows to reduce the additional learning cost because it only transfer the user-view images used in inference without retraining the classification model.

\subsubsection{Road segmentation}
To boost up model performance, we utilize semantically segmented road image into the input composition.
In applying style transfer, our focus lay in bridging the visual differences between street-view and user-view images. This time, however, we make an opposite attempt: what if we emphasize the similarities between the images?
Despite the noticeable dissimilarity between street-views and user-views, there is one thing they share in common - the `\textit{road}'. This road element is relatively stable across both images, even as time passes or weather changes. Thus we guide the model to utilize roads as a reference point with segmented images, 
aiming to better recognize the orientation by finding commonalities between street-view and user-view.

\subsection{Non self-explanatory image classification problem}
For standard supervised classification models, similarities in images with the same label are clearly visible. 
Think of classification task for dogs and cats. Within the training dataset, there are visual commonalities to some extent among the images of dogs, and the same holds true for cats. We term such inherent common features for each label as 'self-explanatory'. 
In our task, however, there are no visual commonalities among images with the same label (e.g., north) because the images are all captured at different locations.
These images cannot be classified by the image itself, leading our classification task to the `non self-explanatory image classification' problem. 
Our main idea to overcome this problem is utilizing context images to determine the angle of user-view. 
To enhance the performance of the orientation detection model,
we placed user-view and context images in various ways to organize input format, and experiments on multiple model structures to find it. Detailed process will be in \S~Evaluation.

\begin{table}[h]
\centering
\begin{tabular}{|c||c|c|c|c|}
 \hline
 & All & Day & Night & Rainy  \\
 \hhline{|=||=|=|=|=|}
 Train image & 148,158 & 148,158 & \-- & \--\\
 \hline
 Test image & 16,533 & 16,533 & \-- & \--\\
 \hline
 Real world image & 410 & 204 & 112 & 94\\
 \hline
\end{tabular}
\caption{Distribution of images by environmental conditions.}
\label{tab:datadistribution}
\end{table}

\begin{figure}[h]
    \centering
    \includegraphics[width=0.65\linewidth]{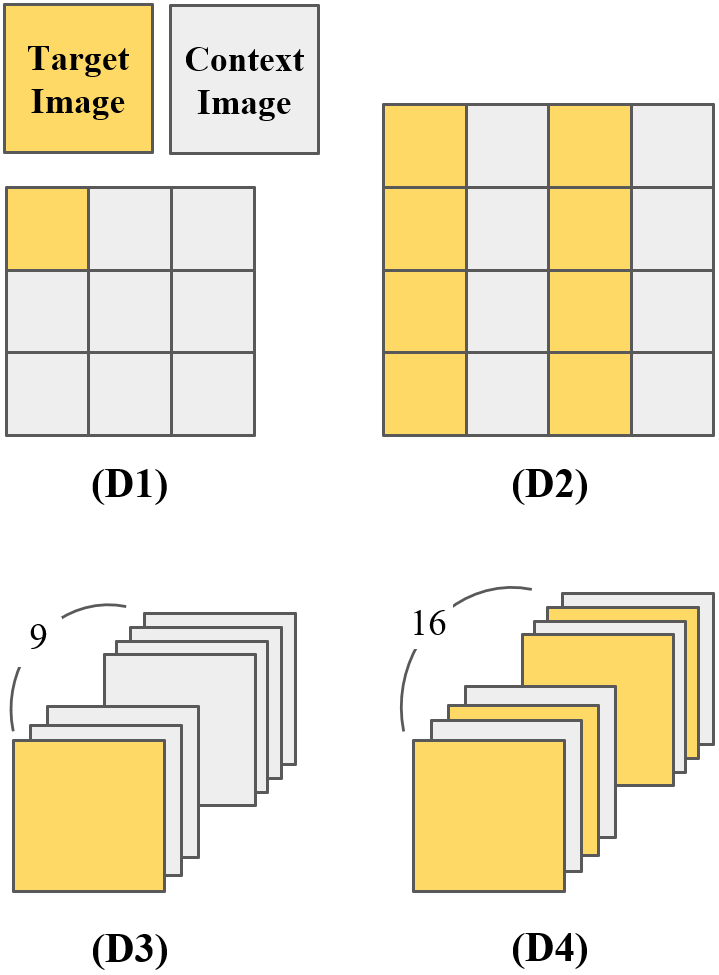}
    \caption{Visualization of the input format. The two main concepts, 2D concatenation (D1, D2) and 3D stacking (D3, D4), are demonstrated.}
    \label{fig:inputformat}
\end{figure}

\section{\prjname}
In this section, we describe the design procedure of \prjname, which involves the process of constructing datasets, designing input format, and artificial transformation.
\subsection{\prjname Dataset}
\subsubsection{Training Dataset}
The optimal approach to detect the orientation where the user is looking at is to construct all the training data using user-view images. However, collecting user-view images labeled with user orientation is challenging. Therefore, we use publicly available street-view rather than using user-view images.
Street-view data can be easily obtained with only GPS coordinates, making it cost-effective. Furthermore, since street-view images are typically collected periodically over time, a variety of images from different time at the similar location can also be easily accessed.
We collected GPS coordinates from a total of 12,064 points on roads. To measure the upper bound accuracy of \prjname, we split these GPS coordinates into a 9:1 ratio for training and testing, and evaluate the accuracy of the street-views themselves. 
For the past-time street-views, within the range of the possible past-time points, we filter out data from the year 2017 onward due to the inferior quality of data before it.

\subsubsection{User-view Dataset}
We collected user-view data to assess the performance of \prjname in real-world environments. The data collection process involved driving a Hyundai Avante MD along the roadways. Specifically, we captured road-travel videos using a GoPro HERO 10 and manually labeled frames extracted from the videos. The GoPro HERO 10 was mounted besides the rearview mirror (approximately 1.3 meters above the ground) and recorded the front view using a 27mm narrow-angle mode.
We gathered a total of 410 front-facing images from user’s perspective to compose the user-view dataset. This consists of images reflecting typical street-view environments (daytime images in clear weather) and images captured in unique situations (daytime images with rain/nighttime images in clear weather). The data distribution for each category can be found in Table ~\ref{tab:datadistribution}. 
The collected user-view images are utilized as test data for model validation, along with eight context images extracted from the most recent street-view at the same GPS coordinate.

\subsection{Input Format Design}
Our goal is to find the orientation of the target image by referring to the context images. The target image is the one we want to classify its orientation (e.g., user-view). Therefore, our approach requires eight context images and at least one target image as the input to the model.
However, typical vision deep learning models are unfamiliar with such input configuration. Thus it is necessary to explore the right way to feed this format of input to the model.
We design the input format with two main concepts and validate them through experiments: 1) 2D concatenation and 2) 3D stacking.
First, to utilize high-performance single image classification models without structural modification, we construct a single square image by concatenating the target and context images. In detail, we test two different concatenation methods: (D1) 3x3 with single target image, (D2) 4x4 with eight copy of target images. 
Second, 
Thus video-processing models are utilized as we believe that they have sufficient potential to gain knowledge from channel-wise image frames. We test two different staking methods: (D3) stacking one target image and eight context images sequentially, (D4) alternately stacking eight duplicated target images and eight context images. A visual representation of these input formats are seen in Figure~\ref{fig:inputformat}. In terms of classification model, we conducted tests on two transformer-based models for each approach - ViT~\cite{dosovitskiy2020image} and SwinV2~\cite{liu2022swin} for 2D concatenation, MViTv2~\cite{li2022mvitv2} and Video Swin-T~\cite{liu2022video} for 3D stacking. 

\begin{figure}
    \centering
    \includegraphics[width=0.8\linewidth]{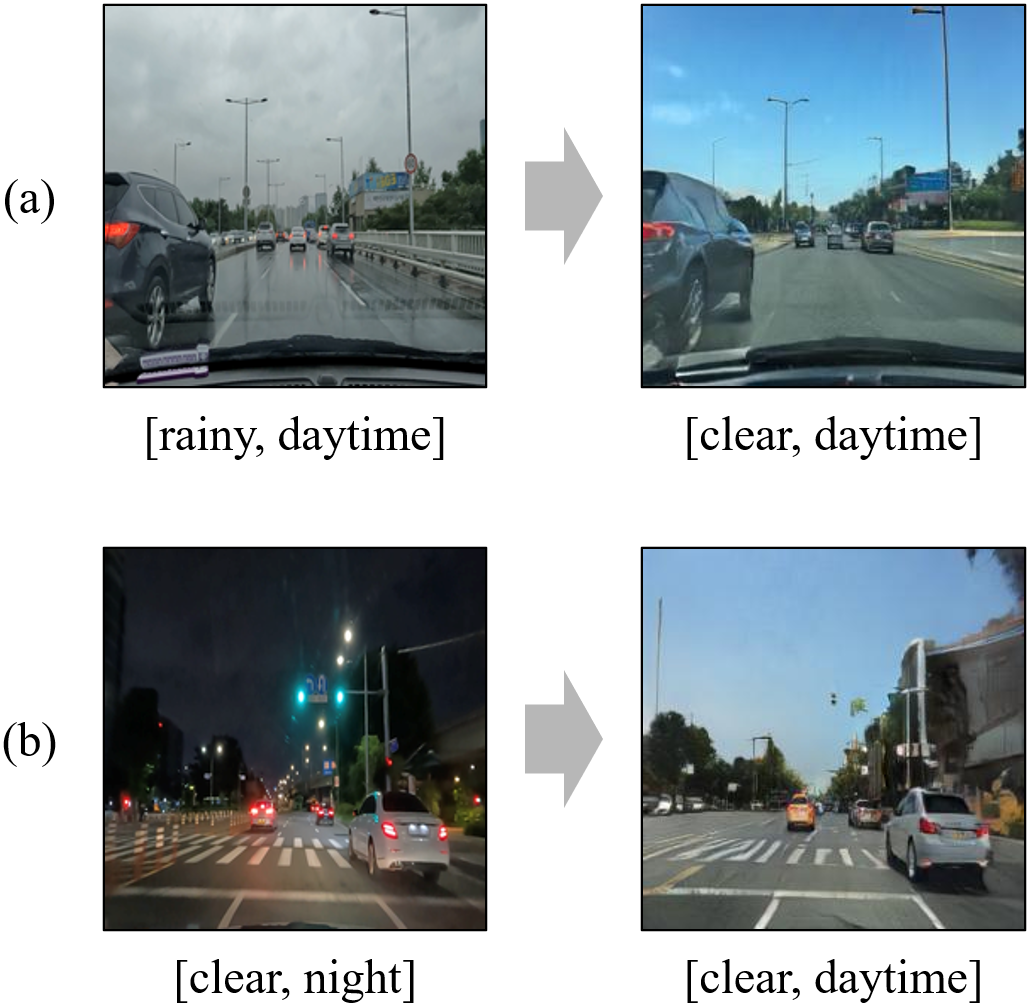}
    \caption{Example of style transfer. Original user-views (left) are style-transferred (right): (a) rainy-to-clear, and (b) night-to-daytime. }
    \label{fig:style_transfer}
\end{figure}

\subsection{Artificial Transformation}
\subsubsection{Style Transfer}

Most of the publicly accessible street-views are captured on clear, sunny days during daytime and thus may not accurately reflect the user's current moment. 
We use style transfer on the user-view to street-view images to reduce the variations between them. As illustrated in Figure~\ref{fig:style_transfer}, [rainy, daytime] images and [clear,night] images are transformed into [clear, daytime] images, in order to ensure the real-world target image works well in a model trained only with street-views. 

We select StarGAN v2~\cite{choi2020stargan} that addresses significant scalability over multiple domains. The user-view street scenes taken in-the-wild are employed as source images, and a randomly selected street-view is utilized as a reference image. The user-view depicting [rainy, daytime] and [clear, night] scenarios are transformed into [clear, daytime] images. The dataset for training StarGAN v2 consists of three different domains: [clear, daytime], [clear, night], [rainy, daytime]. We train the model with combination of several datasets, including SODA10M dataset~\cite{han2021soda10m}, ACDC dataset~\cite{SDV21}, Dark Model Adaptation dataset~\cite{daytime:2:nighttime}, and BDD100K dataset~\cite{yu2020BDD100K}.
As for style transfer, various diffusion-based models have also been recently investigated, so we tested diffusion-based models as well. However, it was not easy to control the model to modify only the desired parts in the images compared to GAN-based models. Since our main goal was to determine the validity of the style transfer approach we proposed, we initially tested it using GAN-based models. We include detailed on the challenges we faced with diffusion-based style transfer in the supplementary materials.


\subsubsection{Road Segmentation (RS)}
To mitigate temporal difference between street-view and user-view, we focused on roads, which are less affected by environmental changes. We aimed for the model to recognize both global features and local road areas. Thus, we used the road-segmented target image as a model input.

\begin{figure}[h]
    \centering
    \includegraphics[width=0.8\linewidth]{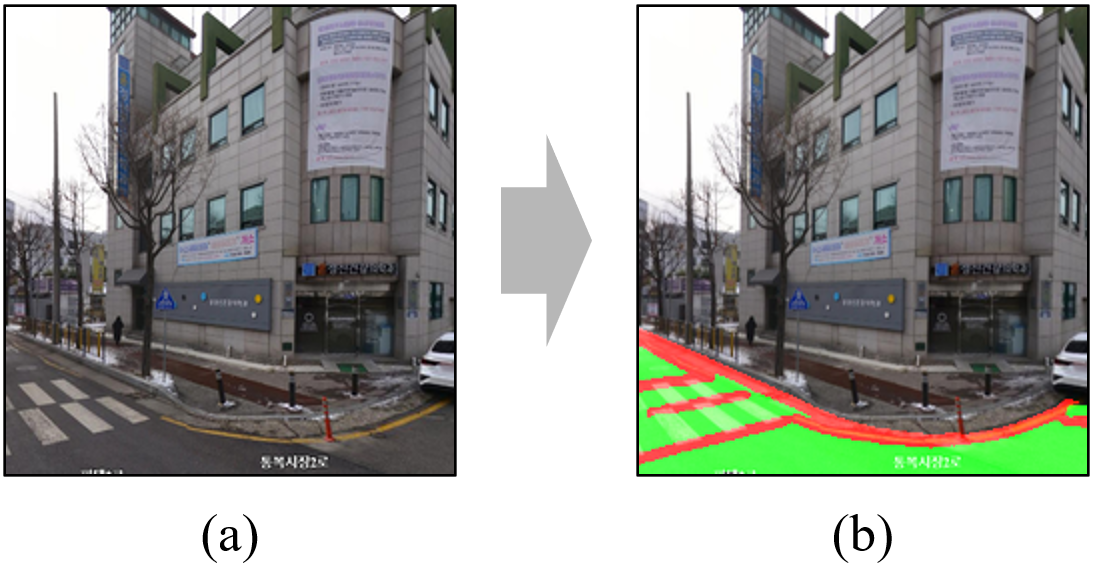}
    \caption{Example of road segmented target image: (a) original RGB target image, and (b) segmented target image.}
    \label{fig:segmentationexample}
\end{figure}

We select YOLOPv2 (Han et al. 2022), which has the state-of-the-art performance on the BDD100K dataset, to segment roads from target images. We utilize a pre-trained model without additional training. In the target images, we mark the roads in green and the dividers in red. Other elements except roads are kept in their original form. Examples can be seen in Figure~\ref{fig:segmentationexample}. 
In the input format using multiple target images (D2, D4), we apply road segmentation to all target images as we described.
\begin{table*}
\small
\centering
\begin{tabular}{|c|c|c|c|c||c||c|c|}
 \hline
 Dim & Index & \multicolumn{2}{c|}{Models}  & Input Size(T x H x W x C) & Upper Bound Acc & Full Data Acc & Daytime Data Acc \\
 \hhline{|=|=|=|=|=||=||=|=|}
 \multirow{6}{*}{2D} & \textit{i} & \multirow{2}{*}{ViT\_B\_16} & D1 & 1x384x384x3 & 87.07\,\% & \textbf{50.49}\,\% & 67.65\,\% \\
 \cline{2-2} 
 & \textit{ii} &  &  D2 & 1x384x384x3 & 88.36\,\% & 42.2\,\% & 61.27\,\% \\
 \cline{2-8}
 &  \textit{iii} & \multirow{2}{*}{ViT\_B\_32} & D1 & 1x384x384x3 & 86.58\,\% & 40.0\,\% & 60.78\,\% \\
 \cline{2-2}
 & \textit{iv} &  & D2 & 1x384x384x3 & 85.24\,\% & 37.32\,\% & 56.86\,\% \\
 \cline{2-8}
 &  \textit{v} & \multirow{2}{*}{Swin\_V2\_B} & D1 & 1x672x672x3 & 81.52\,\% & 12.44\,\% & 18.16\,\% \\
 \cline{2-2}
 & \textit{vi} & & D2 & 1x896x896x3 & 91.45\,\% & 30.24\,\% & 52.45\,\% \\
 \hline
 \multirow{3}{*}{3D} & vii & \multicolumn{1}{c|}{MViT\_V2\_S} & D4 & 16x224x224x3 & 80.09\,\% & 31.22\,\% & 40.2\,\% \\
 \cline{2-8}
 & \textit{viii}  & \multirow{2}{*}{Swin3D\_B} & D3 & 9x224x224x3 & 86.09\,\% & 36.83\,\% & 48.53\,\% \\
 \cline{2-2}
 & \textit{ix} & &  D4 & 16x224x224x3 & 86.31\,\% & \textbf{45.61}\,\% & 61.27\,\% \\
 \hline
\end{tabular}
\caption{Comparative accuracy of ViT-based and Swin-based models across different input formats and dimension.}
\label{tab:experiment result1}
\end{table*}

\begin{table*}
\small
\centering\
\begin{tabular}{|c|c||c||c|c|c|c|}
 \hline
 Index & Models & Upper Bound Acc & Full Data Acc & Daytime Data Acc & Night Data Acc & Rainy Data Acc \\
 \hhline{|=|=||=||=|=|=|=|}
 \textit{a} & \multicolumn{1}{l||}{ViT\_B\_16(D1)} & \multirow{2}{*}{86.58\,\%} & 50.49\,\% & \multirow{2}{*}{67.65\,\%} & 30.36\,\% & 37.23\,\%\\
 \cline{1-1}
 \textit{b} & {with style transfer} & & 58.02\,\% &  & 50.0\,\% & 46.81\,\% \\
 \cline{1-7}
  \textit{c} & \multicolumn{1}{l||}{ViT\_B\_16(D1) + seg} & \multirow{2}{*}{87.44\,\%} & 54.15\,\% & \multirow{2}{*}{71.57\,\%} & 31.25\,\% & 43.62\,\%\\
 \cline{1-1}
 \textit{d} & with style transfer &  & \textbf{63.9}\,\% &  & 58.04\,\% & 54.26\,\%\\
 \cline{1-7}
  \textit{e} & \multicolumn{1}{l||}{Swin3D\_B(D4)} & \multirow{2}{*}{86.31\,\%} & 45.61\,\% & \multirow{2}{*}{61.27\,\%} & 30.36\,\% & 29.79\,\%\\
 \cline{1-1}
 \textit{f} & with style transfer &  & 53.9\,\% &  & 50.89\,\% & 41.49\,\%\\
 \cline{1-7}
 \textit{g} & \multicolumn{1}{l||}{Swin3D\_B(D4) + seg} & \multirow{2}{*}{91.35\,\%} & 46.34\,\% & \multirow{2}{*}{65.59\,\%} & 26.79\,\% & 27.66\,\%\\
 \cline{1-1}
 \textit{h} & with style transfer &  & 49.02\,\% &  & 38.39\,\% & 25.53\,\%\\
 \hline
\end{tabular}
\caption{Comparative accuracy of artificial transformation on MViT and Swin3D using the best input format from Table~\ref{tab:experiment result1}.}
\label{tab:experiment result2}
\end{table*}

\section{Evaluation}
In this section, we describe the experimental results with our implications. Specifically, we focus on two main aspects: 1) input format and model structure, 2) effects of artificial transformation. Then we assess our results compared to the upper-bound accuracy, which is computed on the test dataset composed only of street-view images.


\subsection{Comparison Between Models for 2D\&3D Input}
We used four different types of input formats (D1, D2, D3, and D4) for each model to find which format and model are fit for \prjname. Our experimental results are shown in Table \ref{tab:experiment result1}. We obtained two interpretations from D1 and D2 input of their 2D models.

We observed that the ViT-based model consistently outperformed the SwinV2-based model across all experiments. 
Notably, the SwinV2 model in experiment \textit{v} shows unsatisfactory results compared with others.
We believe that this drop in performance arose due to how the model embeds the data. 
The input data for \prjname consists of a combination of target images and context images. When attempting to extract specific information from such data, it is crucial to first understand the relation between concatenated images. However, in the case of the SwinV2 model, as it's input patch size is very small, it tends to focus on extracting the details of each image rather than understanding globally.
In \prjname, understanding the relationships between concatenated images is more essential than capturing the fine details of the images. As a result, models that excel in general image classification using small patches seem to be unsuitable for \prjname.
The difference between ViT\_B\_16 and ViT\_B\_32 also eventually differs in patch size when embedding images, and the better accuracy of ViT\_B\_16 with larger patches makes our interpretation more valid.

Also, we compared input formats, referring D1 and D2. The differences between experiments \textit{i}, \textit{ii} and \textit{iii}, \textit{iv} shows that 
the accuracy using the D1 were better than those using the D2.
We assume that this difference arises from the size difference of individual image.
Due to the fixed input size of the model, as the number of images consisting the input increases, the size of each individual image decreases, resulting in less information in each image. Consequently, D1, which had fewer images compared to D2, seemed to perform better. In the case of D1, the size of a single image is 128x128, whereas for D2, it is 96x96. For 2-dimensional model, D1 is more suitable, which has a larger individual image size.

From the experimental results of the 3D model using D3 and D4, we compared MViT\_V2\_S and Swin3D\_B in terms of accuracy. Referring to Table \ref{tab:experiment result1}, MViT achieved an accuracy of 31.22\%, while Swin3D\_B achieved higher accuracy with 45.61\%. Although both models are designed to utilize features from various layers, there is a difference in their number of parameters. Swin3D has 88.0M parameters, while MViT having 34.5M. In our belief, the number of parameters of MViT seems to be insufficient to effectively handle our task.

\begin{figure*}
    \centering
    \includegraphics[width=0.84\textwidth]{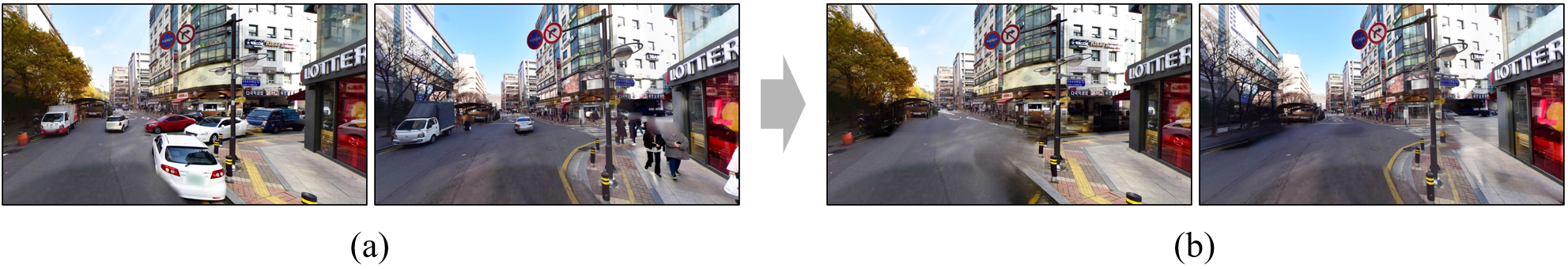}
    \caption{Example of removing dynamic objects. (a) are the original versions at different time, and (b) are the results of removal}
    \label{fig:img_compare}
\end{figure*}

\subsection{Artificial Transformation}
With the best performing model and the corresponding input format from Table~\ref{tab:experiment result1}, we verify the effect of artificial transformation techniques to overcome the discrepancy between the street-views and the user-views. The results are shown in Table~\ref{tab:experiment result2}.
In comparison to the results without road segmentation (A, E), the outcomes with road segmentation applied (C, G) are notably higher, suggesting the effectiveness of our segmentation approach. On the Full Data, ViT exhibited the most significant improvement with an enhancement of 3.66\%.
 

Style transfer approach demonstrated superior results compared to road segmentation. With the ViT model, there was an improvement of X\%, while the Swin3D model exhibited a Y\% improvement. These results validate the effectiveness of style transfer. Similarly, in road segmentation, the improvement was more significant on ViT than on Swin3D.
Based on our analysis, Swin3D appears to be more sensitive to noise induced by style transfer than ViT, likely due to its larger input image size. Although StarGAN\_V2 effectively conducts style transfer, there is perceptible loss in details that induces noise during the transformation process. We estimate that employing images smaller than 224x224 would be more beneficial when applying style transfer.


Using both approaches simultaneously did not always work better, but the maximum performance was the highest. We achieved 63.9\% with ViT\_B\_16(D1) in validation experiments using all real-world images. With this model, we achieved 73.07\% of it's upper-bound(87.44\%).


\section{Discussion and Limitation}

\subsection{Extension to Indoor Setting}
We demonstrated the potential of \prjname in outdoor driving situation, yet it is not limited. 
\prjname is applicable even in complex indoor spaces such as large shopping centers and international airports. While GPS often has limited functionality in these indoor environments, we can still determine users location using indoor map. There are also scattered cues, such as stores, that can be used to infer orientation.
In this case, the street-view approach may not be applicable due to the unavailability of GPS-based localization. However, it is possible to obtain a visual context of user's surroundings through CCTV, allowing users to determine their orientation and navigate within such spaces.


\subsection{Dealing with Dynamic Objects}
There is a temporal difference (e.g., appearance and viewpoint) between two images of the same location taken at different times. \prjname attempts to mitigate the discrepancy in weather and illumination by applying deep learning techniques. However, many dynamic objects such as vehicles and pedestrians are present in the images, which can interfere with determining precise orientation. 
As such, we have considered adopting an approach that utilizes image segmentation and inpainting techniques~\cite{10.1145/3411764.3445688,10024513} to remove these dynamic objects and reconstruct the images. ~\ref{fig:img_compare} shows examples where this process has been applied. While we have not implemented this method in our study, there are still rooms for further investigation. We expect that future improvements with these techniques could make \prjname more robust in handling gap between images.
\subsection{Limitations}
\textbf{Limited in-the-wild test data coverage}
We proposed several methods to make \prjname robust on environmental changes. However, due to limitations in data collection and its cost, we were unable to accumulate enough real-world test data to verify DeepCompass's capability across various situation. In this paper, we have demonstrated the potential by confirming reasonably effective performance in a certain time of day (daytime/night) on certain weather condition (clear/rainy).
Furthermore, it is essential to take camera specifications into consideration, such as its types and angles, to apply \prjname in practice. In the real-world, users employ various vehicles and camera devices (e.g., smartphones, dashcams). Therefore, apart from the current fixed collecting environment, it is necessary to collect data from a wider range of devices and vehicles. We aim to leave the exploration of various other situations for future work.

\noindent
\textbf{Vulnerability in GPS localization error}
\prjname is a orientation detection method which requires accurate GPS coordinates, implying that proper GPS localization is essential to obtain the corresponding street-view. In other words, \prjname relies on GPS localization, and the extent of GPS errors is closely related to the drop in the performance of \prjname. To overcome such issue, we have considered not only using a single location measured by the GPS but also utilizing street-views of nearby coordinates within a certain range. We believe this will allow us to accurately infer orientation even with minor GPS error. In future research, we aim to address and improve upon this vulnerability.

  
\section{Conclusion}
\prjname tackles the problem of asynchronous location/orientation identification. This problem leads to minor yet frequently occurring inconveniences such as incorrect navigation guidance at the beginning of driving. It leverages street-view images from different points in time to identify the user's orientation without magnetometers. We use several deep learning techniques to minimize the discrepancy between street-view and real-world image, and have developed \prjname, which can detect orientation in real-world scenarios using models trained on street-views.


\bibliography{aaai24}



\section*{Supplementary material}

# ***Supplementary Material***
# DeepCompass: AI-driven Location-Orientation Synchronization for Navigating Platforms

| Dataset description | Same-time (%) | Diff-time (%) |
|---|---|---|
| Same-time data | 87.08 | 26.92 |
| Diff-time data | 72.25 | 61.75 |
| All-time data | 81.58 | 63.0 |

Table 1: Performance of models trained on same-time and different-time data (All-time refers to combined data of same-time and diff-time.

## Composition of Training Data

**Same-time vs Different-time period**   During the design process of DeepCompass, we conducted an experiment to determine how to effectively utilize street-view (i.e. context image) as training data to identify the orientation of the user-view (i.e., target image). In real-life situations, the time period between the context image and the target image is always different, yet it is possible to formulate those data under the same time period in the lab setting. However, it is hard to ensure that such a setting could reflect reality. Therefore, we compared the model's accuracy when it was trained only with data under the same time period (hereafter denoted as same-time) and when it was trained with data under different time periods (hereafter denoted as diff-time). Figure 1 presents how to structure the experimental data and Table 1 shows the result. We used Swin3D model with the ImageNet 1k checkpoint. We experiment DeepCompass on street-view images because we were interested in verifying how much same-time setting reflects the real world. As a result, the model trained on same-time images showed an advantage only within that specific time window (achieved an accuracy of 87.08%), yet the accuracy on diff-time data achieved 26.92% accuracy. In addition, the model trained on combined data of same-time and diff-time did not significantly improve the performance (an increase of 1.25%). Since the same-time was not able to reflect reality, we decide to use diff-time data for this study.

**Policy for Selecting Different-time Data**   When using context images of diff-time along with target images, various differences arise due to their temporal disparities. In addition, it is difficult to identify the orientation when the positions of context and target images are significantly different. Specifically, position changes between context and target images lead to variations in the perceived background even though their orientations are the same (see Figure 1). To mitigate this problem, we constrained the coordinate differences between the target image and context images. The Constraint is as follows:

$$|t\_lat - c\_lat| + |t\_long - c\_long| < 0.0001$$

In the equation above, $t$ and $c$ denote the target and context, and $lat$ and $long$ denote latitude and longitude, respectively (e.g., $c\_lat$ corresponds to context latitude). An example of a range that satisfies the following constraints is shown in Figure 2.

| Dataset description | diff-time (%) |
|---|---|
| 45-degree | 61.75 |
| 30-degree | 50.33 |

Table 2: Performance comparison of models based on Angle differences

## Extended Experiment

In this section, we additionally describe our early results that are excluded in the manuscript.

**Angle Diversity**   Basically, DeepCompass is designed to classify the target's orientation in eight different orientations with the context images composed with fixed degrees (i.e. 45 degree). We thought that training with context images composed of various angles might work better. We tried training the model using randomly selected eight images out of the twelve images divided with 30-degree intervals to create the context images. The results of the experiment are presented in Table 2. Contrary to our expectations, the performance was not better than fixed 45-degree context data. We think that learning with various angles might have been too difficult, or perhaps the significant positional changes caused by variations in angles made it challenging for the model.

**Angle Labeling**   In deep learning research, labeling is an important issue that affects the performance of a model. This is especially crucial for DeepCompass, as it deals with a problem called 'non-self-explanatory image classification', since there are no visually common characteristics among labels. To address this, we proposed two different approaches, in-between orientation and matching orientation. 1) In-between orientation method (corresponding to

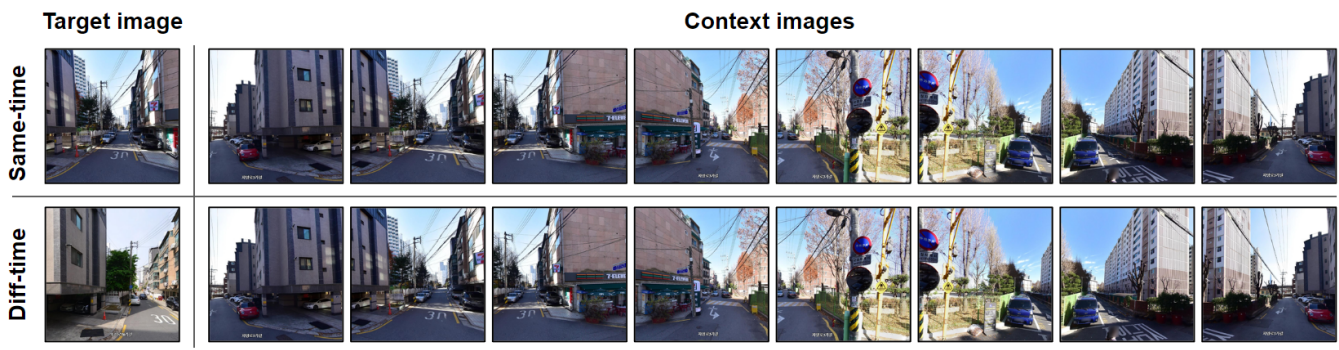

Figure 1: Target images from the same time and different time point as the context images (The labels of the two target images are the same)

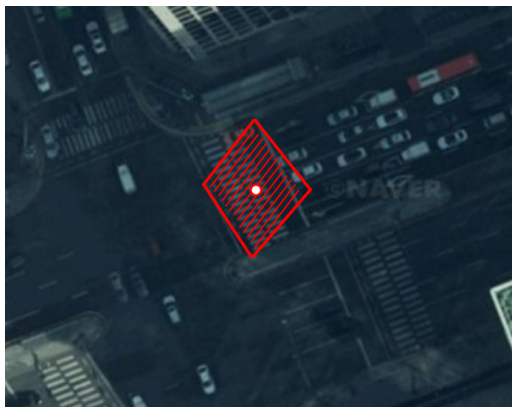

Figure 2: The range of latitude and longitude differences that satisfy the constraints in actual aerial images (where the white dot represents the target's location)

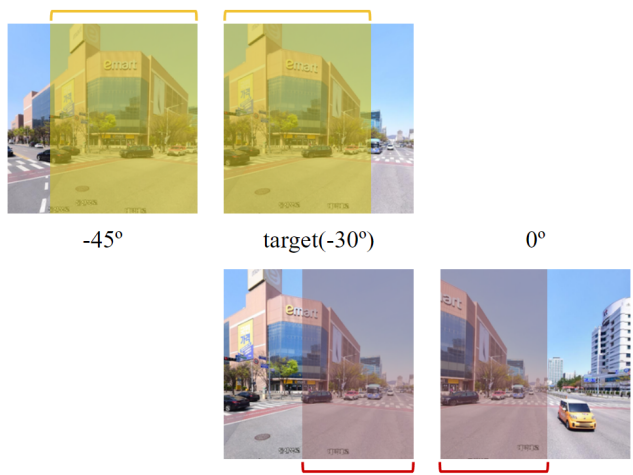

Figure 3: Adjacent context between the target image and neighboring context images

| Dataset description | Street view data(%) | In the wild (%) |
| --- | --- | --- |
| Between Orientation | 91.82 | 62.25 |
| Matching Orientation | 86.84 | 54.19 |

Table 3: Performance comparison of models based on data labeling differences

Figure 4a) determines the orientation by identifying where the target image should be positioned between two context images that are cut at regular angle-interval and arranged in sequence, 2) Matching orientation method (corresponding to Figure 4b) determines the direction by finding which image from the ordered context images is most similar to a particular image. It could seem that there is not much difference (in terms of the 22.5-degree shift). However, as shown in Table 3, the accuracy difference is quite significant (accuracy of 4.98% for street-view and 8.06% for in-the-wild experiment). This result highlights the significance of determining labels in the Non self-explanatory image classification task.

## Extended Discussion

**Diffusion Model-Based Style Transfer** In this study, we used style transfer to enhance the performance for real-world test. We aim to transform real-world view to street-view like images (i.e., to the clear daylight). With the recent advancement of diffusion-based image generation models, the traditional GAN-based models have been largely replaced in many aspects. Therefore, we conducted an experiment with a Diffusion-based model called CycleDiffusion (Wu and la Torre 2022). The results of the experiment are depicted in Figure 5, Our goal was to modify the lighting conditions (day/night) or weather (clear/rainy) without significantly altering objects that can be used for orientation. However, the results showed excessive changes in objects such as trees appearing or changing shape, and even the background mountain disappearing. Therefore, in this study, we used GAN-based model (StarGAN) instead of a diffusion-based model as our choice for the style transfer model.

**In The Wild Data Collection** The test data for real-world was collected through our own efforts. As a result, a significant cost was incurred in collecting real-world data. Due

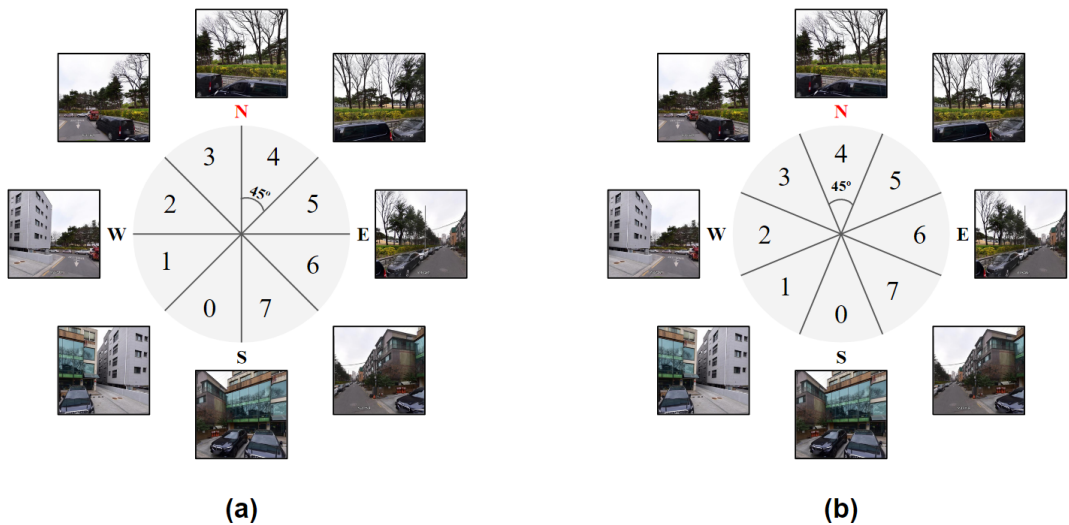

Figure 4: in-between orientation(a) and matching orientation(b) method

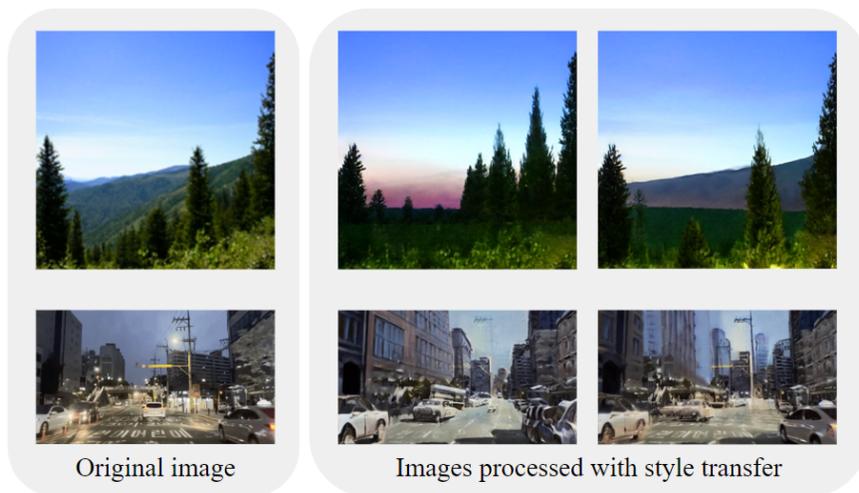

Figure 5: Examples of style transfer.

to the difficulties of collecting data in various environmental settings, we were unable to conduct tests in rare weather conditions or in diverse geographical regions. With the utilization of automated specialized equipment or vehicles designed for street view photography, it would be possible to gather more accurate testing data. We also expect that building on such collected data enables us to conduct experiments in diverse weather conditions and environmental changes. This would lead to the more robust DeepCompass.



\end{document}